\pdfoutput=1

\documentclass[11pt]{article}

\usepackage[final]{EMNLP2023}

\usepackage{times}
\usepackage{latexsym}

\usepackage{amsmath}
\usepackage{amssymb}

\usepackage{booktabs}

\usepackage{listings}
\usepackage{enumitem}

\usepackage{tcolorbox}

\usepackage{adjustbox}
\usepackage{lscape}
\usepackage{rotating}

\usepackage[T1]{fontenc}

\usepackage[utf8]{inputenc}

\usepackage{microtype}

\usepackage{inconsolata}

\newtcolorbox{mybox}[1]{colback=cyan!5!white,colframe=cyan!75!black,fonttitle=\bfseries,title=#1}

\usepackage{todonotes}
\makeatletter
\newcommand*\iftodonotes{\if@todonotes@disabled\expandafter\@secondoftwo\else\expandafter\@firstoftwo\fi}  %
\makeatother

\title{Large GPT-like Models are Bad Babies: A Closer Look at the Relationship between Linguistic Competence and Psycholinguistic Measures}

\author{Julius Steuer\quad\quad Marius Mosbach\quad\quad Dietrich Klakow \\
        Department of Language Science and Technology \\ Saarland University \\
         \texttt{\{jsteuer,mmosbach,dietrich.klakow\}@lsv.uni-saarland.de}
        }

\begin{document}
\maketitle
\begin{abstract}

Research on the cognitive plausibility of language models (LMs) has so far mostly concentrated on modelling psycholinguistic response variables such as reading times, gaze durations and N400/P600 EEG signals, while mostly leaving out the dimension of what \citet{mahowald2023dissociating} described as formal and functional linguistic competence, and developmental plausibility. We address this gap by training a series of GPT-like language models of different sizes on the strict version of the BabyLM pretraining corpus, evaluating on the challenge tasks (BLiMP, GLUE, MSGS) and an additional reading time prediction task. We find a positive correlation between LM size and performance on all three challenge tasks, with different preferences for model width and depth in each of the tasks. In contrast, a negative correlation was found between LM size and reading time fit of linear mixed-effects models using LM surprisal as a predictor, with the second-smallest LM achieving the largest log-likelihood reduction  over a baseline model without surprisal. This suggests that modelling processing effort \emph{and} linguistic competence may require an approach different from training GPT-like LMs on a developmentally plausible corpus.

\end{abstract}

\section{Introduction}

\begin{figure}[!t]
    \centering
    \includegraphics[width=0.5\textwidth]{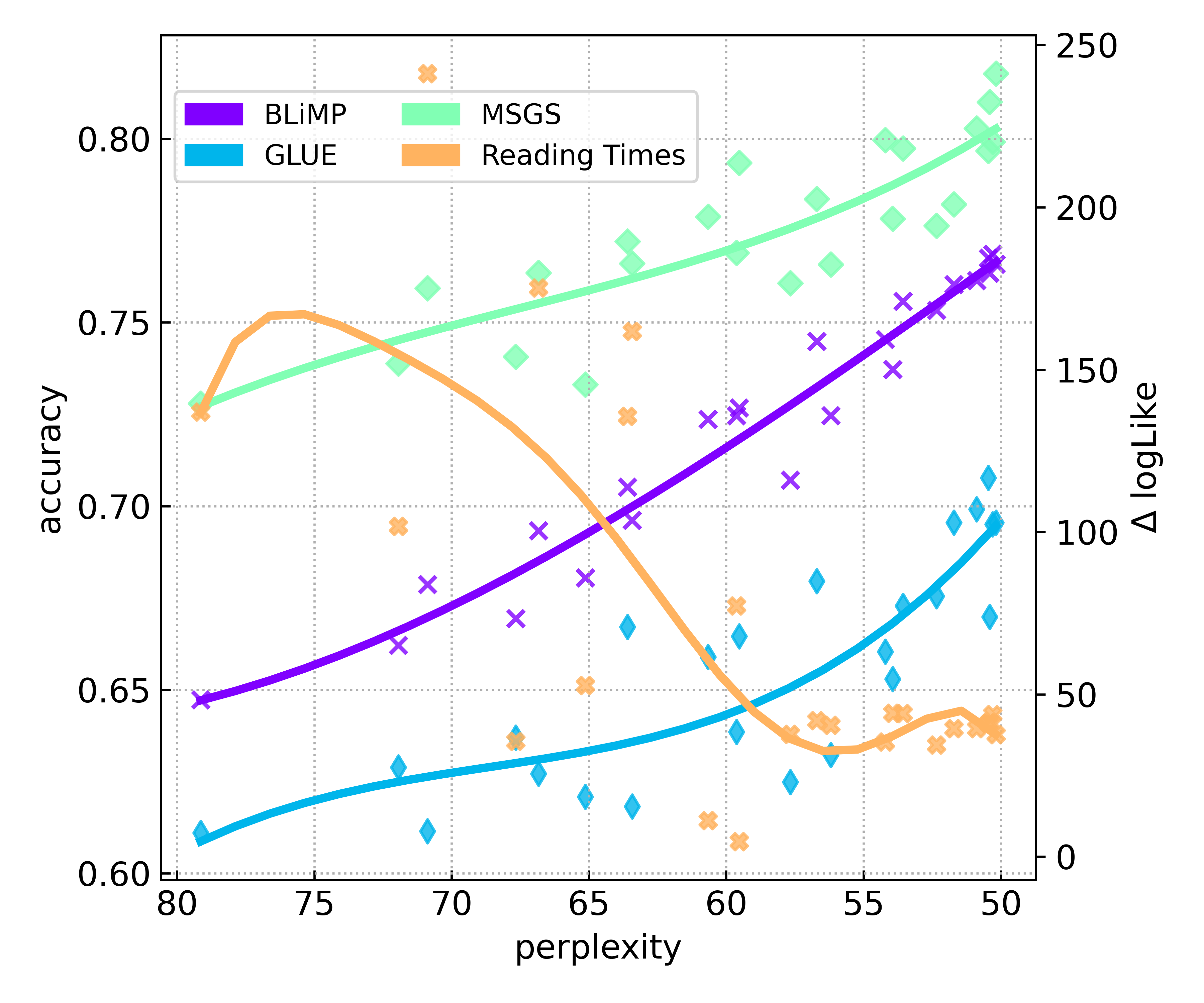}
    \caption{Our results show that LM performance on the BabyLM challenge tasks is negatively correlated with perplexity on the development set of the BabyLM corpus (lower perplexity leads to higher performance). In contrast, a \emph{positive} correlation (Spearman's $\rho=0.4784$, $p<0.05$) was found between LM perplexity and the fit of LM surprisal to self-paced reading times from the Natural Stories corpus \citep{futrell_natural_2021} in terms of the difference in log-likelihood between a basline linear mixed-effects model and a model using LM surprisal as a predictor. Lines were fitted with 3 (challenge tasks) or 6 (reading times) degrees of freedom to the LMs' average performance on the task. See Section \ref{sec:results} for detailed results.}
    \label{fig:perf_by_pp}
\end{figure}

In recent years several approaches have been taken to test LMs for cognitive plausibility. This is usually done by using output probabilities of the LM as a predictor for a model's preference towards certain linguistic structures \citep{roark_deriving_2009, wilcox_predictive_2020}. Another strain of research uses the output probabilities as a correlate of psycholinguistic measures, e.g., N400 and P600 EEG signals (\citealp{heilbron_tracking_2019} and recently \citealp{li_decomposition_2023}) and (self-paced) reading times \citep{fernandez-monsalve-etal-2012-lexical}. A natural question that arises is whether cognitive plausibility should be attributed to the model architecture itself, or to the training regime in combination with the training dataset. Little research has been done on the actual neurological plausibility of large LMs (LLMs), but \citet{schrimpf_neural_2021} showed that the architecture of BERT-like models is already plausible for the next word prediction task before training: model predictions with only the language modelling head trained are already predictive of human brain activity during reading \emph{and} correlate well with the predictions of the fully trained model. In contrast, no correlation between brain activity and model predictions was found for models trained on GLUE \citep{wang_glue_2019}, a natural language understanding (NLU) benchmark. This finding may mirror an underlying difference in language processing between \emph{formal} and \emph{functional linguistic competence} as introduced by \citet{mahowald2023dissociating}:

\paragraph{Formal linguistic competence} is defined as the "capacity required to produce and comprehend a given language, i.e., the ability to distinguish grammatically correct from incorrect formations, based either on "knowledge of and flexible use of linguistic rules" or "non-rule-like statistical regularities" \citep{mahowald2023dissociating}. An example for the former mechanism would be the regular formation of past tense verbs in English (\emph{look}:\emph{looked}), and for the latter the formation of irregular or ablauting past tense verbs (\emph{go}:\emph{went},\emph{tread}:\emph{trod}).  

\paragraph{Functional linguistic competence} is defined as "non-language-specific cognitive functions that are required when we use language in real-world circumstances" \citep{mahowald2023dissociating} , i.e., the ability to perform cognitive tasks \emph{with} language. GLUE is an example for a benchmark that test this dimension of linguistic competence, with some if its tasks (CoLA \citep{warstadt2019neural}) also testing for aspects of \emph{formal linguistic competence}.
\\
\\
The dichotomy between formal and functional linguistic competence can be understood in terms of Wittgenstein's definition of the meaning of a word as its use in a language (\citet{wittgenstein53german}, §43). The debate on whether statistical learners (i.e. LMs) can learn the meaning of a linguistic unit (word, phrase, text, etc.) in Wittgenstein's sense is still ongoing, with much division between positions that strongly deny that LMs can have such a property \citep{bender-koller-2020-climbing} and positions that advocate that they might have it, e.g., under the condition that the LM's predictions are grounded in extralinguistic reality \citep{bisk-etal-2020-experience}. Our study does not attempt to find arguments in favour of either position, 
but to study the implications of this dichotomy for the paradigm of cognitive modelling.

As stated earlier, the output probabilities of LMs lie at the basis of the application of LMs to cognitive language modelling, usually in the form of a probability distribution over a vocabulary of word forms given either surrounding words (masked language modelling) or preceding words (causal language modelling). Evidence for the use of surprisal (a word's negative logarithmic probability in context) instead of the actual probablity comes from logarithmic effects of contextual probabilities on processing difficulty \citep{shain_large-scale_2022}. Another approach is to evaluate the output probabilities of a LM over a number of classes that may or may not apply to the input sequence, usually after fine-tuning the LM. The reliance of research in this direction on the output probabilities of LMs has already been criticized from multiple sides. There is a growing body of evidence that the performance of a LM in the typical language modelling task, next word prediction, and measures of formal linguistic competence are not correlated. \citet{hu_systematic_2020} found no correlation between LM perplexity and measures of formal linguistic competence, while \citet{huang_surprisal_2023} argue that LM surprisal should not be assumed to be a good predictor of psycholinguistic measures of processing difficulty that require more than just lexical information. This lack of correlation with psycholinguistic measures becomes more prominent with the increasing size of LMs \citep{oh_why_2022}, and especially so in extreme cases of human processing difficulty: \citet{arehalli_syntactic_2022} showed that surprisal from LSTM-based LMs underestimates garden-path effects on reading times, while successfully predicting reading times for most non-garden-path sentences. This finding has been corroborated for transformer-based LMs such as GPT-2 \citep{jurayj_garden-path_2022} and BERT \cite{irwin_bert_2023}.

\section{BabyLM}

The BabyLM challenge \citep{warstadt-et-al-2023-babylm} introduces a novel constraint to cognitively plausible language modelling by limiting the token budget for LM pretraining to 100 million (100M) tokens, roughly the same amount of tokens a 13-year old child has seen during language acquisition \citep{gilkerson_mapping_2017}. While the focus of the challenge is on the pretraining procedure, the evaluation pipeline consists of the BLiMP \citep{warstadt_blimp_2020},  MSGS \cite{warstadt_learning_2020} and GLUE benchmarks, each of which aims to test for a specific dimension of linguistic competence.

\paragraph{BLiMP}
BLiMP tests for \textit{formal linguistic competence} by comparing model predictions at a critical word in pairs of grammatically acceptable and unacceptable sentences, with the sentence pair only differing with respect to a single feature, e.g., whether a determiner agrees with its antecedent in gender or not. A model succeeds at the task if it assigns a higher probability to the critical word in the acceptable sentence.

\paragraph{GLUE}
GLUE is a benchmark that requires fine-tuning\footnote{During fine-tuning, we train all parameters of the pretrained LM as well as a randomly initialized classifier on top of the LM.} of the LM. It tests for a wide range of NLU problems, e.g., question answering, natural language inference and linguistic acceptability judgements, and hence can be regarded as a proxy for the \emph{functional linguistic competence} of a LM.

\paragraph{MSGS}
MSGS is a benchmark of binary classification tasks that tests whether a LM prefers \emph{surface generalizations} over \emph{syntactic generalization} by first fine-tuning on data consistent with both types of generalization. At inference time, items are consistent with only one type, potentially revealing a bias towards either generalization type.
\\
\\
\noindent Previous studies mainly provided insights into the relationship of pretraining token budget and measures of formal and functional linguistic competence. \citet{zhang-etal-2021-need} showed that encoder-only LMs already perform well on formal tasks such as BLiMP at a budget of 10-100M tokens, while requiring substantially larger token budgets to perform well on functional tasks such as GLUE. While this research established correlations for pretraining token budgets, similar relationships for \emph{model size} at a fixed token budget have not yet been investigated. This study is dedicated to finding a relationship between model size and performance on these tasks, while simultaneously addressing the dimension of \emph{processing effort}, which is not covered by the challenge tasks. This is done using the \textbf{strict} version of the BabyLM corpus, mainly because there is evidence that the fit with psycholinguistic measures profits from token budgets far larger than the 100M tokens in the corpus \citep{oh_transformer-based_2023}.%
However, we also implicity evaluate on models that are trained on token budgets of 10M tokens, corresponding rather to the \textbf{strict-small} track in Section \ref{sec:rts_exp}.

\section{Research questions}

The starting point of our work is \citet{zhang-etal-2021-need}'s finding of an earlier saturation effect (in terms of pretraining tokens) for BLiMP as opposed to (Super)GLUE. If performance on BLiMP is already close to the optimum after pretraining for 100M tokens, we suspect that a model with relatively small capacity is sufficient to reliably learn the required syntactic and semantic features. In contrast, the larger pretraining token budget and model size needed for GLUE should also require a model with higher capacity. 

Studies on reading time prediction generally use causal LMs trained on a next-word prediction task instead of masked LMs \citep{oh_why_2022, arehalli_syntactic_2022, jurayj_garden-path_2022} because of their closer similarity to human language processing. Although masked LMs such as BERT show some word order effects \cite{papadimitriou_when_2022} and even garden-path effects \cite{irwin_bert_2023}, they are cognitively implausible in the sense that they process all words in a sequence simultaneously when predicting a word at a masked position, rather than processing language sequentially. This \emph{autoregressive} property mirrors human language processing, and is therefore  desirable in studies with the primary goal of modelling human reading behaviour. We therefore employ decoder-only, GPT-like LMs \citep{radford_language_nodate} in our study, i.e., we want to answer the following research questions:\looseness-1

\begin{mybox}{Research question A}
    Are GPT-like models cognitively plausible in the sense that they are able to acquire (a degree of) formal and functional linguistic competence, while being also predictive of human processing effort?
\end{mybox}

\begin{mybox}{Research question B}
    Can such LMs be trained on the same data as a child has available during language acquisition (100M tokens)?
\end{mybox}

\section{Previous work}

\paragraph{Do we need transformers for cognitive plausibility?}

Despite promising findings by \citet{hosseini_neural_2021}, it has yet to be determined whether transformers, and decoder-only transformer LMs in particular, are cognitively plausible in the sense that they are data-efficient enough to acquire human-like\footnote{Here, we do not use "human-like" to imply human-level performance, but rather that the model is \emph{subject to similar processing constraints} as a human.} linguistic competence. Indeed, there are results that seem to partially contradict the necessity of LLMs with wide context windows in order for a model to exhibit human-like processing behaviour. \citet{kuribayashi_context_2022} showed that \emph{reducing} context length of LLMs improves the fit of a linear mixed-effects model (LME) on gaze durations, with surprisal from a bigram GPT-2 model as a predictor yielding the largest log-likelihood reduction over the baseline model. \citet{wilcox_predictive_2020} failed to identify a relationship between psychometric predictive power ($\Delta$ log-likelihood) and syntactic generalization, concluding that different models are needed for modelling human processing effort versus syntactic generalization. %

\paragraph{Linguistic competence vs. psycholinguistic measures}

It has long been clear that LM capacity, and subsequently LM perplexity, does not necessarily correlate with human-likeness \citep{kuribayashi_lower_2021}. LLMs such as GPT-3 in particular were found to have considerable disadvantages when it comes to predicting psycholinguistic measures from their next-word predictions: \citet{oh_why_2022} found an inverse relationship between both perplexity and LLM capacity, versus fit to human reading times. The authors of this study hypothesize that this is because transformers have access to the full sequence context, and are trained on large enough corpora to make use of the information that they contain. This relationship between model perplexity and reading times is however not intrinsic to transformer-based LMs: \citet{hu_systematic_2020} found a similar relationship for LSTM LMs, though small GPT-like models have an advantage over recurrent models.\looseness-1 

The impact of LM size on linguistic competence was investigated by \citet{eldan_tinystories_2023}, who found that relatively small GPT2-like models ($<$10M parameters) manage to produce fluent English and can be trained on relatively small corpora with a reduced vocabulary. Their study also shows that the relationship still holds for small models, while also identifying trade-offs between model width (hidden size) and depth (number of decoder layers).\looseness-1 

As for training dataset size, \citet{oh_transformer-based_2023} found that surprisal from transformer-based LLMs gives the best fit to reading times at about 2B train tokens, across a wide range of model sizes. The corpus used in their study is very large (300B tokens), allowing for extensive training of a model without repeating any data. Reaching the same number of update steps with the much smaller BabyLM corpus would require training for multiple epochs.

\paragraph{Single- vs. multi-epoch training}
Since the BabyLM training data is substantially smaller than the 2B tokens suggested by \citet{oh_transformer-based_2023}, training our models in a multi-epoch setting cannot be avoided. Previous research has shown that repeating the training data can have adverse effects: \citet{xue_repeat_2023} compared single-epoch vs. multi-epoch training in a limited data setting and show that multi-epoch training leads to overfitting, with little performance being gained after the first epoch. They also find that regularization can only partially alleviate the overfitting problem, with dropout having the largest effect.
\noindent
Not having to repeat the training data is advantageous for downstream tasks and psycholinguistic modelling, if a certain amount of training data is available: \citet{oh_transformer-based_2023} found that reading time fit deteriorates after 2B tokens over a wide range of model sizes. However, it is not clear if repeating the training data would lead to an even stronger deterioration. If the corpus is substantially smaller than 2B tokens, repeating the training data could have a different effect, especially if the optimum of the reading time fit depends on the availability of the 2B tokens.

\section{Methodology}

\label{sec:methodology}

\begin{figure*}[!t]
    \centering
    \includegraphics[width=\textwidth]{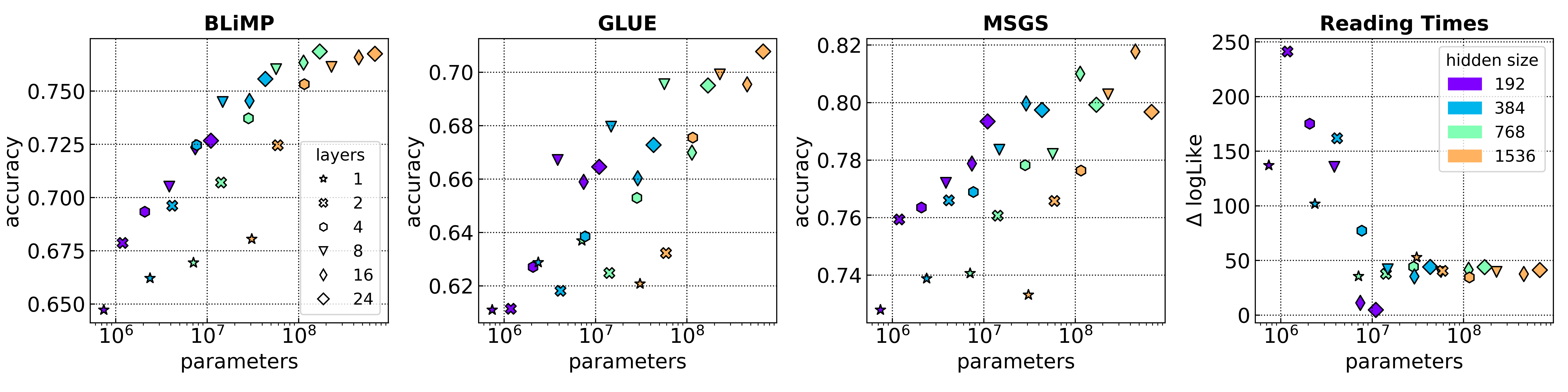}
    \caption{Task performance by model size (higher numbers are better). Baselines can be found in Appendix \ref{sec:appendix_b}.}
    \label{fig:all_tasks}
\end{figure*}

\paragraph{Modelling}

We use the OPT architecture by \citet{zhang_opt_2022} with a language modelling head for pretraining. Following our intuition that BLiMP should require much smaller model sizes than MSGS and GLUE, we train a series of OPT models of different sizes, varying only model width (hidden size) and model depth (number of decoder layers). In total we train 24 models varying over 4 hidden sizes $l_{hidden} \in \{192, 384, 768, 1536\}$ and 6 numbers of decoder layers ($l_{decoder} \in \{1, 2, 4, 8, 16, 24\}$). We also adjust the dimension of the feedforward layers such that the size of the output vector $l_{forward} = 3 \times l_{hidden}$. Table \ref{tab:opt_models} in Appendix \ref{sec:appendix_a} shows the resulting model sizes. The models and all code for pretraining are implemented with PyTorch \cite{pytorch} and HuggingFace transformers \cite{wolf-etal-2020-transformers}, starting from their implementation of OPT. We also trained a new tokenizer on the training set of the BabyLM corpus, using the same vocabulary size $|V| = 50272$ as the original OPT tokenizer. We report all results as averages over 3 random seeds (see Appendix \ref{sec:appendix_b} for full results and standard error).

\paragraph{Training}

Following the Shortformer pipeline \citep{press_shortformer_2021}, each model is trained for one epoch with an initial sequence length of 64, followed by 4 epochs with the full sequence length of 256. The full sequence length of 256 was chosen as a compromise between the relatively short test items in the challenge tasks (up to 128 tokens) 
In order to ensure that the model generalizes to longer sequences we use ALiBI \cite{press_train_2022} instead of learned positional embeddings. This also ensures that our models generalize to the longer sequences in the Natural Stories corpus. We trained each model on a A100 GPU with 40 GB VRAM and an effective batch size of 128, using gradient accumulation for models that could not fit the full batch size. We used AdamW \citep{adamw} as our optimizer with an initial learning rate of $0.001$ and weight decay of $0.001$ with 2000 linear warm-up steps. We use a dropout of 0.1 following the default HuggingFace transformers parameters for OPT.

\paragraph{Pretraining experiments}

We also experimented with changes to the pretraining regime. We trained models on multiple permutations of the training dataset: ordering sequences according to length (number of words), word length (number of characters), sequence-level perplexity from a 3-gram LM trained on the same data, and different orderings of the subcorpora as in \citet{mueller_how_2023}. None of these approaches resulted in significant performance gains in terms of perplexity and performance on the challenge tasks over a baseline model trained on the concatenated BabyLM corpus with subsequent shuffling of the sequences.

\paragraph{Evaluation}

We evaluated all models on the downstream tasks of the BabyLM challenge. While these three tasks test for the linguistic competence of a model, they do not quantify the cognitive effort associated with language processing. We therefore also evaluate all models on a reading time prediction task. For each model, we calculated surprisal on the items of the Natural Stories Corpus \cite{futrell_natural_2021}. This corpus was chosen because its domain is close to at least one of the BabyLM subcorpora (Children's Stories). We fitted linear mixed-effects (LME) models with random intercepts for subject, word and item (the id of the story); surprisal, word frequency, word length and sentence position as predictors and log-normalized reading times as the response variable. The exact formula is
\\

\lstset{basicstyle=\footnotesize}

\begin{lstlisting}[breaklines]
  log(reading_time) ~ 
    word_surprisal + len(word)
    + log(word_frequency) + position
    + (1|word) + (1|subject) + (1|item)
\end{lstlisting}

\noindent
For the reading time analysis we report the difference in log-likelihood between the models with surprisal as a predictor over a baseline model with only the control predictors. For all other tasks we report accuracy. 

\paragraph{Code}

We used the evaluation code provided by the organizers of the BabyLM challenge\footnote{\url{https://github.com/BabyLM/evaluation-pipeline}}, with some modifications to load custom models. The evaluation pipeline is based on the LM-Eval framework by \citet{eval-harness}. Fine-tuning on GLUE and MSGS was done with the default hyperparameter settings, but we reduced the number of fine-tuning epochs to 3 as we did not observe any improvements after 3 epochs. The LME models were fitted using the lmerTest R library \cite{lmerTest} via the pymer4 Python package \cite{pymer4}. The code to pretrain and evaluate all models is publicly available on GitHub\footnote{\url{https://github.com/uds-lsv/babylm}}. The model with the highest BLiMP accuracy and detailed results for the LME models are made available at the same location, alongside instructions on how to run the training and evaluation pipelines.

\section{Results}

\label{sec:results}

\begin{figure*}[!t]
    \centering
    \includegraphics[width=\textwidth]{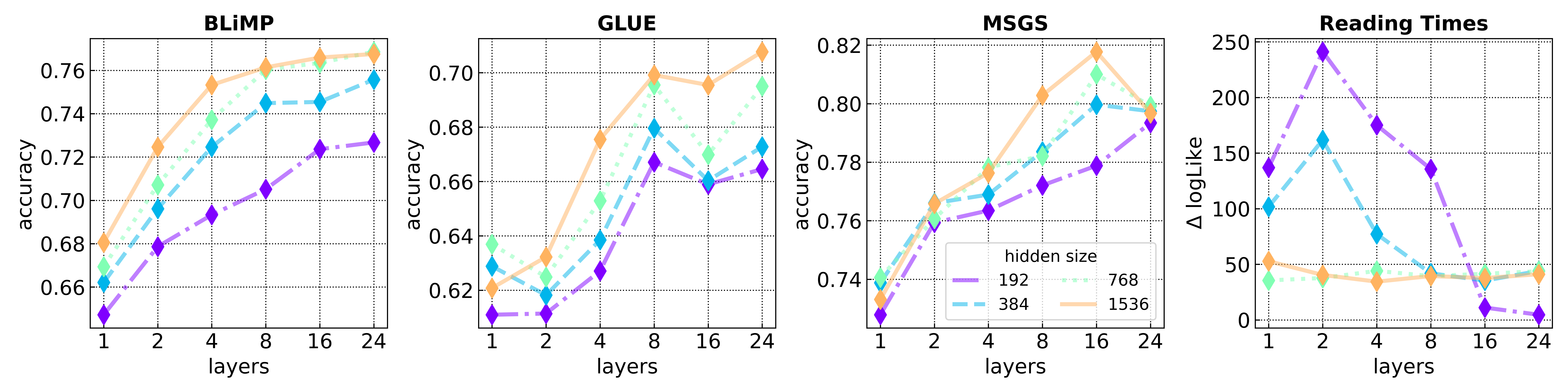}
    \caption{Task performance by hidden size, number of layers and task.}
    \label{fig:radar_plots}
\end{figure*}

\paragraph{Fine-tuning GLUE}

 Fine-tuning on GLUE was overall very unstable and often failed to outperform the baseline. This was mainly due to the one-size-fits-all approach to the fine-tuning hyperparameters; we repeated several more fine-tuning runs with different hyperparameter settings on some of the GLUE tasks, and found that, e.g., RTE profited from a longer warm-up period (which is in line with the findings of \citet{mosbach2021stability} for BERT-like models), but most other sub-tasks fine-tuned with the same hyperparameters showed a drop in performance. While we could have optimized hyperparameters for all sub-tasks, the main objective of the BabyLM challenge is to improve the pretraining part of the NLP pipeline. Thus, we decided to fine-tune with the default hyperparameters, only adjusting the number of epochs as we found that the fine-tuning runs already converged after a few epochs.

\paragraph{Model size}

Figure \ref{fig:all_tasks} shows the relationship between model size and task performance: While GLUE (Spearman's $\rho=0.7739$, $p<1^{-4}$) and MSGS ($\rho=0.7148$, $p<1^{-4}$) performance scales with model size, BLiMP performance plateaus after reaching a model size of about 50M parameters ($\rho=0.8835$, $p<1^{-4}$). In contrast, reading time fit was negatively correlated with model size ($\rho=-51.39$, $p<0.05$). All correlations are statistically significant with $p<1^{-4}$. No single model performed best on all three challenge tasks, with large differences in the size of the best model. Figure \ref{fig:perf_by_pp} shows that similar correlations hold for model perplexity and task performance (BLiMP: $\rho=-0.9765$, $p<1^{-4}$, GLUE: $\rho=-0.8287$, $p<1^{-4}$, MSGS: $\rho=-0.8661$, $p<1^{-4}$); negative correlations mean that lower perplexity leads to higher performance. We found strong positive correlations (pictured in Figure \ref{fig:blimp_vs_others} in Appendix \ref{sec:appendix_b}) between performance on the challenge tasks (BLiMP and GLUE ($\rho=0.8784$), BLiMP and MSGS ($\rho=0.9182$) and GLUE and MSGS ($\rho=0.815$) generally with $p<1^{-4}$).

\paragraph{Model width vs. depth}

While BLiMP performance  was not found to be strongly correlated with either the number of decoder layers or hidden size, GLUE and MSGS showed some variability based on the number of layers. For GLUE the only configuration that showed a monotonic improvement in performance was a hidden size of 1536, with models with more decoder layers achieving higher accuracy in this setting. For MSGS we observed a drop in performance for the models with 24 decoder layers at the largest hidden sizes (384, 768). Overall, the effect of hidden size and number of layers was minor when compared to overall model size. In contrast, the best fit on the reading time data was achieved with the second smallest model with only 2 decoder layers and a hidden size of 192.  Figure \ref{fig:radar_plots} illustrates this trend: for the challenge tasks, performance increases with the number of layers (though not monotonically), whereas $\Delta$ log-likelihood of the LME models decreases with the number of layers at $l_{hidden} = 192$ and, to a lesser extent, at $l_{hidden} = 384$, while deeper models with more decoder layers and larger hidden sizes perform considerably worse.

\paragraph{Possible confounds}

The reading time analysis suffers from several potential confounding factors: Firstly, the domain of the training data differs considerably from the data in the Natural Stories corpus. While the training data also contains some longer texts (Wikipedia, Children's Stories), most of the corpora are more representative of spoken language (Open Subtitles, BNC Spoken, CHILDES). In addition, most sequences are relatively short, with a median sequence length of 8 in the Open Subtitles corpus, which accounts for $>$50\% of the training data. This is considerably less than the median sequence length of 22 in the Natural Stories corpus.
\noindent
Another confounding factor might be the difference in exposure to language data of the model and that of the participants of the original reading time study. \citet{futrell_natural_2021} do not provide demographic data of their participants, but since data collection was done via Amazon Mechanical Turk we can safely assume that the mean age of the participants was higher than 13, meaning that they were exposed to considerably more language data than the 100M tokens in the BabyLM corpus. Although a recent study by \citet{oh_transformer-based_2023} showed that reading time fit (in terms of $\Delta$ log-likelihood) from transformer models still profits from pretraining data multiple orders of magnitude larger than our corpus, with an optimum at 2B tokens, this is partially alleviated in this study by the multiple-epoch training regime, totalling about 500M tokens seen by each of our models. Since \citet{oh_transformer-based_2023} found that training on more \emph{unseen} tokens after reaching the optimum leads to a quick deterioration of reading time fit proportional to model size, it is unclear what impact repeating the training data would have on the reading time fit. 

\section{Reading time prediction in a multi-epoch setting}

\label{sec:rts_exp}

\paragraph{Experimental setup}

\begin{figure}[!t]
    \centering
    \includegraphics[width=0.5\textwidth]{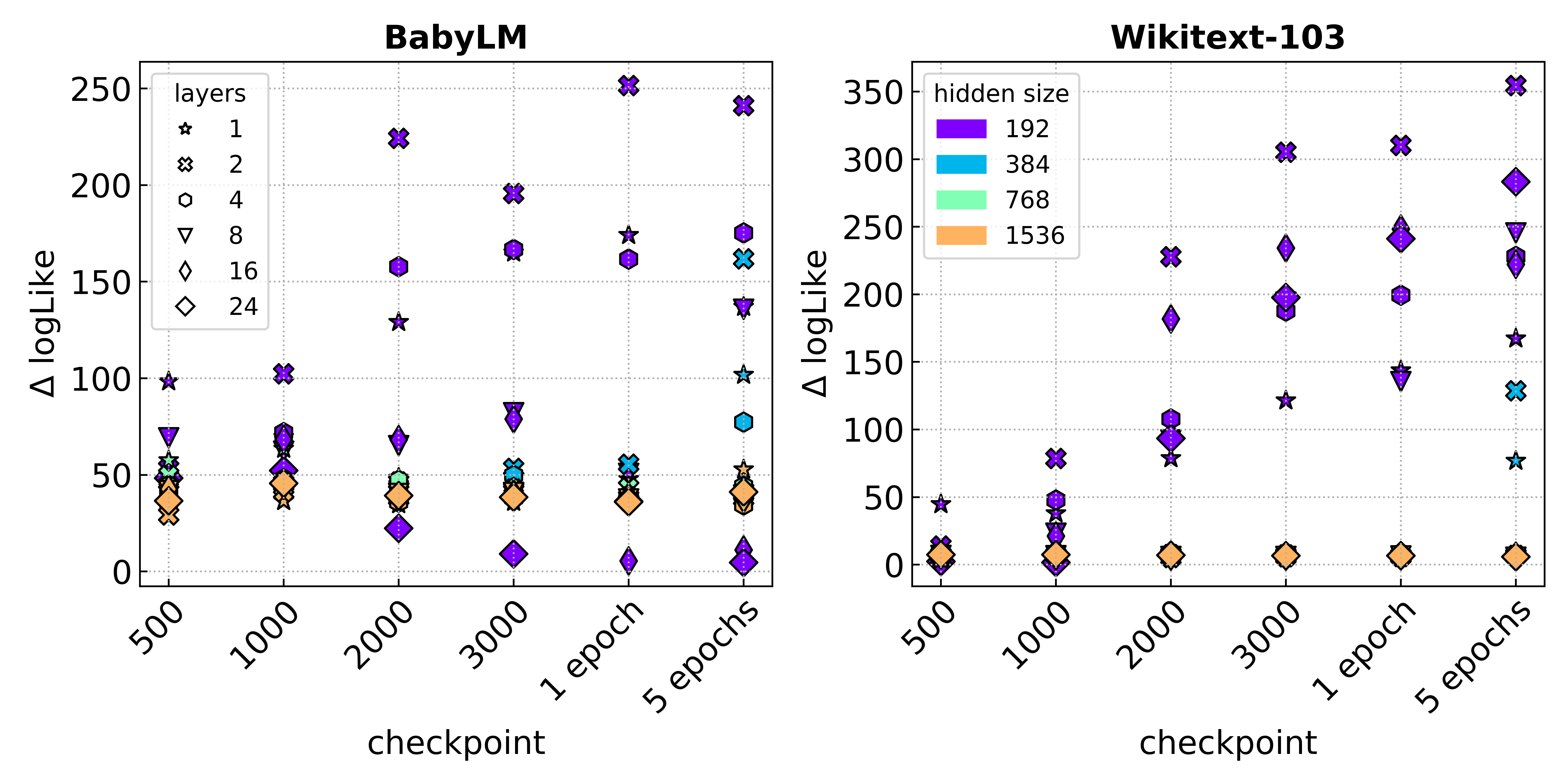}
    \caption{Reading time fit in terms of $\Delta$ log-likelihood over a base model without surprisal as a predictor, on the BabyLM and Wikitext-103 data after 500, 1000, 2000 and 3000 training steps (1/8, 1/4, 1/2 and 3/4 of an epoch) and 1 and 5 epochs.}
    \label{fig:rts_more_steps}
\end{figure}

In order to evaluate whether the negative correlation is an artifact of the domain mismatch between the BabyLM corpus and the items in the Natural Stories corpus or the repetition of the training data before reaching the optimal token budget, we conduct two additional experiments: First, we retrain all models on the BabyLM corpus for a single epoch, saving intermediate checkpoints at 100, 500, 1000, 2000 and 3000 training steps. Then, we use the intermediate models to fit LME models to the reading time data, using the same formula as given in Section \ref{sec:methodology}. Second, we replicate these experiments on Wikitext-103, a corpus of similar size that does not have the same limitations of the BabyLM corpus (i.e. an average sequence length and a domain closer to the Natural Stories corpus). The models trained on Wikitext-103 serve as a control for the experiments on the BabyLM corpus and were not included in the final submission. Since the results indicate that larger models yield a worse reading time fit, we restrict the experiment to small models (1-4 layers, all hidden sizes) and larger models with the smallest and largest hidden size (192 and 1536). The models are trained with the same hyperparameter settings as the original models, but sequence length is not reduced in the first epoch.

\paragraph{Results}
Figure \ref{fig:rts_more_steps} shows a somewhat different picture for the models trained on Wikitext-103, with reading time fit of smaller models increasing over the whole pretraining procedure, while models with $l_{hidden} > 192$ almost never improve over the baseline model. In contrast, the reading time fit of the LMs trained on the BabyLM data improves significantly over the baseline for shallower models ($<$ 2 decoder layers), while staying roughly constant for deeper and wider models (16, 24 decoder layers). However, the relationship between the number of training steps and reading time fit is not monotonic, with a slight decrease after training for 4 more epochs for the best model.
\noindent
While the models trained on the Wikitext-103 dataset yield a better fit to reading times in terms of $\Delta$ log-likelihood, the basic finding on the BabyLM data is corroborated: exposing a transformer model to multiple repetitions of the training data before reaching the optimal token budget does not lead to a decrease in reading time fit, but also does not improve over the single epoch setting in a meaningful way. The results also show that the improved reading time fit for $l_{hidden} = 192$ cannot be attributed to smaller model size alone, as the deepest model with that hidden size, 24*192 shows an improved fit over the baseline, while 1*384, a model with a comparable number of parameters, but a larger hidden size, does not. In conclusion, we did not find a degradation of reading time fit when repeating the training data, with similar effects of LM size on reading time fit for Wikitext-103 and the BabyLM corpus (see Table \ref{tab:detailed_rts_results} in Appendix \ref{sec:appendix_rts} for Spearman's $\rho$'s and p-values). We also found The BabyLM corpus to be advantageous for this task in the sense that -- in contrast to Wikitext-103 -- reading time fit from all models improved over the baseline LME model.

\section{Discussion}

\paragraph{Correlation between BLiMP, GLUE \& MSGS}

\begin{figure}[!t]
    \centering
    \includegraphics[width=0.45\textwidth]{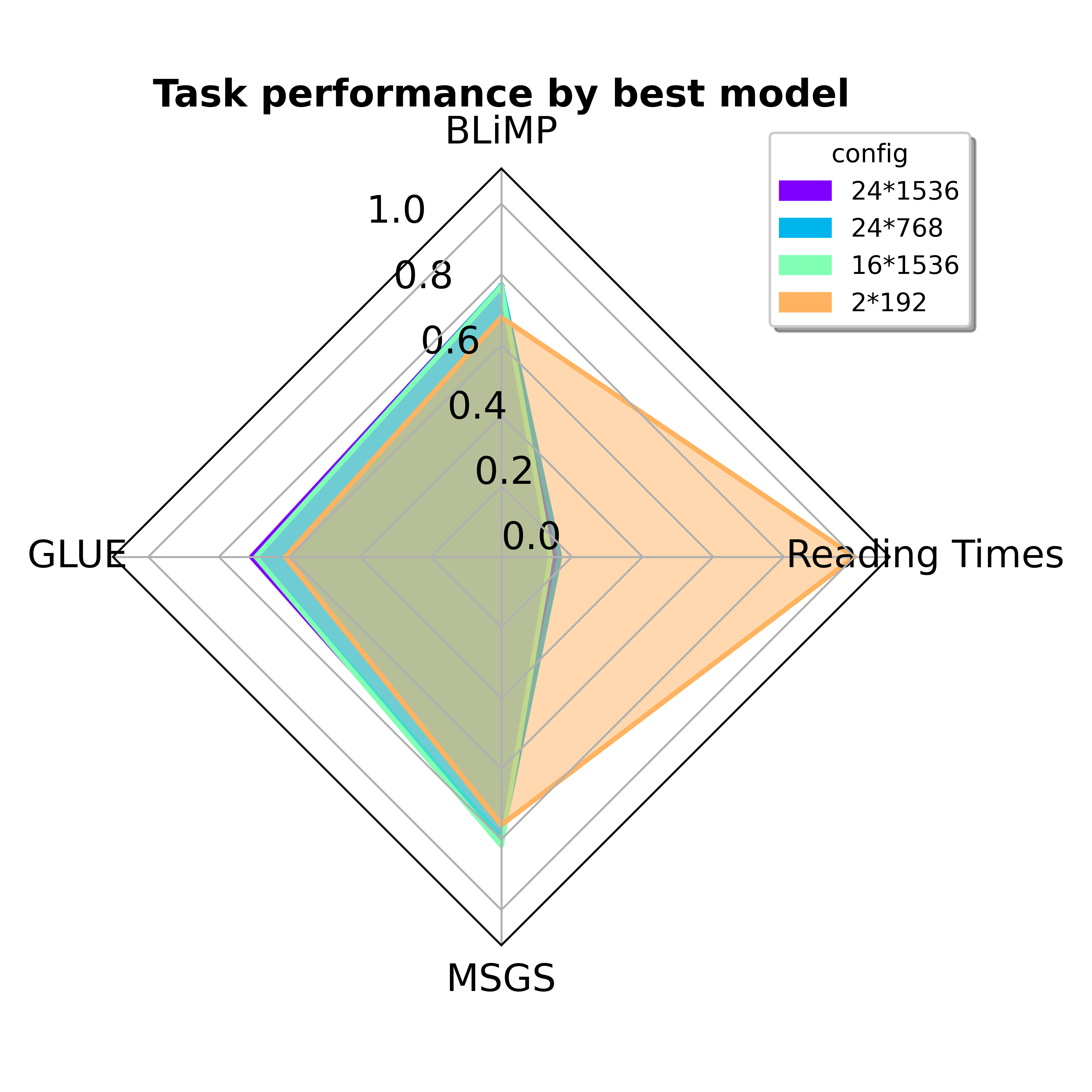}
    \caption{Performance of the best models by task. Reading times $\Delta$ log-likelihoods are normalized in the interval $[0,1].$}
    \label{fig:radar_plot_tasks}
\end{figure}

The experiments presented in Section \ref{sec:results} provide evidence for a correlation between LM performance on BLiMP, GLUE and MSGS tasks when pretraining on the BabyLM corpus. This correlation is in accordance with established effects of training dataset size (\citealp{zhang-etal-2021-need}), and interactions of train corpus size and model capacity (\citealp{eldan_tinystories_2023}, \citealp{kaplan_scaling_2020}). However, no single model achieves the highest score on all three tasks: BLiMP shows diminishing returns for model sizes larger than 50M tokens, while the best model on MSGS (16*1536) is substantially smaller than the best model on GLUE (24*1536). This discrepancy between the best model on the BabyLM challenge tasks and on the reading times prediction task is illustrated by Figure \ref{fig:radar_plot_tasks}. The correlation between BLiMP/MSGS and GLUE may be an artifact of the sub-optimal fine-tuning on GLUE, failing to outperform the baseline model. It cannot be ruled out that the results would change when determining the optimal hyperparameters for each sub-task individually. However, even if the correlation were an artifact of the pretraining data, the findings of a negative correlation between model size and reading time fit would still hold. 

\paragraph{Cognitive plausibility of GPT-like models}

\noindent
The best fit on self-paced reading times from the Natural Stories corpus was obtained with the second-smallest model, with models with $l_{hidden} > 192$ only slightly improving over the baseline. The second suite of experiments in Section \ref{sec:rts_exp} confirms that this is not solely caused by the multi-epoch training regime necessitated by the small token budget. The reason for the mismatch between measures of cognitive plausibility (reading times) and measures of formal (BLiMP, MSGS) and functional linguistic competence (GLUE) is rooted in the interaction of pretraining regime and model size: While it is feasible to train a model that performs comparatively well on all four tasks on a budget of 100M tokens, the sweet spot for model size and dataset size is reached much earlier for the reading time prediction task than for the BabyLM challenge tasks. This problem could easily be resolved by using one model when modelling reading times (or any other psycholinguistic measure), and another model when either of the forms of linguistic competence is the aim. This might be a valid and promising approach in a situation where the understanding of the research object does not depend on the connectedness of its experimental analoga. In the case of our research object -- the human language faculty -- it may not be necessary to find a single analogon that accounts for all its components, but since we \emph{know} that the human language faculty is part of a unified cognitive system (with specialized sub-units) performing the tasks which the modern language modelling pipeline of pretraining and fine-tuning splits up into individual modules, it would be worthwile to move in the direction of a unified approach that accounts for both forms of linguistic competence and empirical evidence of processing effort. This could be achieved through adjustments to the pretraining regime (in terms of data, modelling objective etc.), as suggested by the BabyLM challenge, or through adjustments to the model architecture. 

\paragraph{Size of transformer models} The results of the reading time prediction study on the BabyLM corpus indicate that it in fact has an \emph{advantage} over Wikitext-103, although the LMs trained on the latter achieve larger $\Delta$ log-likelihoods on average: Since the largest models fail to improve over the baseline model if trained on Wikitext-103, it is possible that some properties of the language in the BabyLM corpus facilitate the learning mechanism that actuates the correlation of LM surprisal and reading times. The reason for the worse fit of surprisal from the larger models may be that both Wikitext-103 and the BabyLM corpus are not large enough to induce the learning bias needed to give good predictions of reading times in larger models, with Figure \ref{fig:rts_more_steps} showing that the results on the BabyLM corpus are much less stable than on Wikitext-103 and the improvements over the baseline much less sharply linear. In summary, our results lead to the following answers to our research questions:

\begin{mybox}{Result: Research question A}
    GPT-like LMs can be cognitively plausible and display formal and functional linguistic competence, although not both at the same time...
\end{mybox}
\begin{mybox}{Result: Research question B}
    ...under the constraint of a developmentally plausible training dataset. 
\end{mybox}
\vspace{0.05cm}

\section{Conclusion}

Our study highlights the challenges of training a LM that performs well on tasks requiring some degree of formal and functional linguistic competence as defined by \citet{mahowald2023dissociating}, while also being predictive of the psycholinguistic measure of reading times. We find that small, shallow models of less than 5M parameters yield the best fit to the psycholinguistic measure, while performance on BLiMP, GLUE and MSGS improves with increasing model size, although to a different degree for each of the tasks. This has implications for research on cognitively or developmentally plausible models of human language processing: in the case of a small, domain-specific training corpus it is not feasible to pretrain an LLM that displays formal linguistic competence and performs well on a reading time prediction tasks, a conclusion also drawn by \citet{connell-understanding-2022}. Consequently, research in this direction has concentrated on fine-tuning pretrained LLMs on domain-specific data, e.g., \citet{skrjanec_broy_demberg_2023}. A promising approach to a unified architecture could be relegating special tasks (such as classifying a sequence as in GLUE) to adapters \citep{adapters}, sub-networks within a pretrained LM. This approach is common in multilingual language modelling \citep{pfeiffer_lifting_2022, alabi-etal-2022-adapting}, where its success is partially attributed to its ability to separate general linguistic knowledge from language-specific information. A similar modelling decision may be necessary for cognitively plausible language models.

\section*{Acknowledgements}
The authors thank Iza \v{S}kranjec for helping with the training and interpretation of LME models, and Vagrant Gautam, Michael Hahn, Benedict Schneider, Iza \v{S}kranjec and for their helpful comments. This research was supported by the Deutsche Forschungsgemeinschaft (DFG, German Research Foundation), Project ID 232722074 -- SFB 1102.

\section*{Limitations}

The results of the paper mainly hold for decoder-only transformer LMs. While these LMs are closer to human language processing in the sense that they process language incrementally, this has some disadvantages for reading time predictions, since humans do not attribute equal importance to each word, skipping some words in the process, and typically integrate words from the left- \emph{and} right-hand context of a fixated word. While the first point can be addressed by explicitly modelling skipping behaviour \citep{hahn-keller-2016-modeling}, the second could require a solution closer to masked language models.

A second limitation is the focus on self-paced reading time as the psycholinguistic response variable. Since the setup of self-paced reading studies, with the participants observing a single word at a time, distorts the natural reading process, the measure itself may be not that cognitively plausible. This could be addressed by repeating the experiments on corpora from eye-tracking studies such as the Dundee corpus \citep{dundee_corpus}. There is evidence that much larger models than those
tested in the current study still improve the fit to total reading times in less restricted experimental settings \citep{de-varda-marelli-2023-scaling}. The latter study also shows that the fit to psycholinguistic measures varies over languages and writing systems.

Another option is modelling brain activity patterns directly by predicting N400 and P600 EEG signals, which have the additional advantage of providing a means of decomposing LM surprisal without the proxy of linguistic structure, as shown by \citet{li_decomposition_2023}.

\section*{Ethics Statement}

The authors foresee no ethical concerns about the work presented in the paper.

\bibliography{anthology,custom,references}
\bibliographystyle{acl_natbib}

\appendix
\onecolumn

\section{OPT models}
\label{sec:appendix_a}

\def\arraystretch{1.5}%
\begin{table}[h!]
    \centering
        \begin{tabular}{c|c|c}
        \hline
        $l_{decoder}$ & $l_{hidden}$ & \textbf{Parameters (non-embedding)} \\
        \hline \hline
        1  & 192 & 0.74  \\
        2  & 192 & 1.19 \\
        4  & 192 & 2.07\\
        8  & 192 & 3.85 \\
        16 & 192 & 7.41 \\
        24 & 192 & 10.9 \\
        \hline
        1  & 384 & 2.37  \\
        2  & 384 & 4.14 \\
        4  & 384 & 7.69 \\
        8  & 384 & 14.79 \\
        16 & 384 & 28.99\\
        24 & 384 & 43.18 \\
        \hline
        1  & 768 & 7.09  \\
        2  & 768 & 14.18 \\
        4  & 768 & 28.35 \\
        8  & 768 & 56.70\\
        16 & 768 & 113.41\\
        24 & 768 & 170.11 \\
        \hline
        1  & 1536 & 30.69 \\
        2  & 1536 & 59.00 \\
        4  & 1536 & 115.69 \\
        8  & 1536 & 229.01 \\
        16 & 1536 & 455.67\\
        24 & 1536 & 682.32 \\
        \hline
    \end{tabular}
    \caption{OPT models sizes in million parameters by hidden size and number of decoder layers. The number of parameters does not include the embedding table, which is always of the size $l_{emb} \times |V| = 768 \times 50272 = 38.608.896$, as in OPT-128m.}

    \label{tab:opt_models}
\end{table}

\newpage
\section{Validation perplexity}
\label{sec:appendix_pp}
\begin{figure}[!h]
    \centering
    \includegraphics[width=0.8\textwidth]{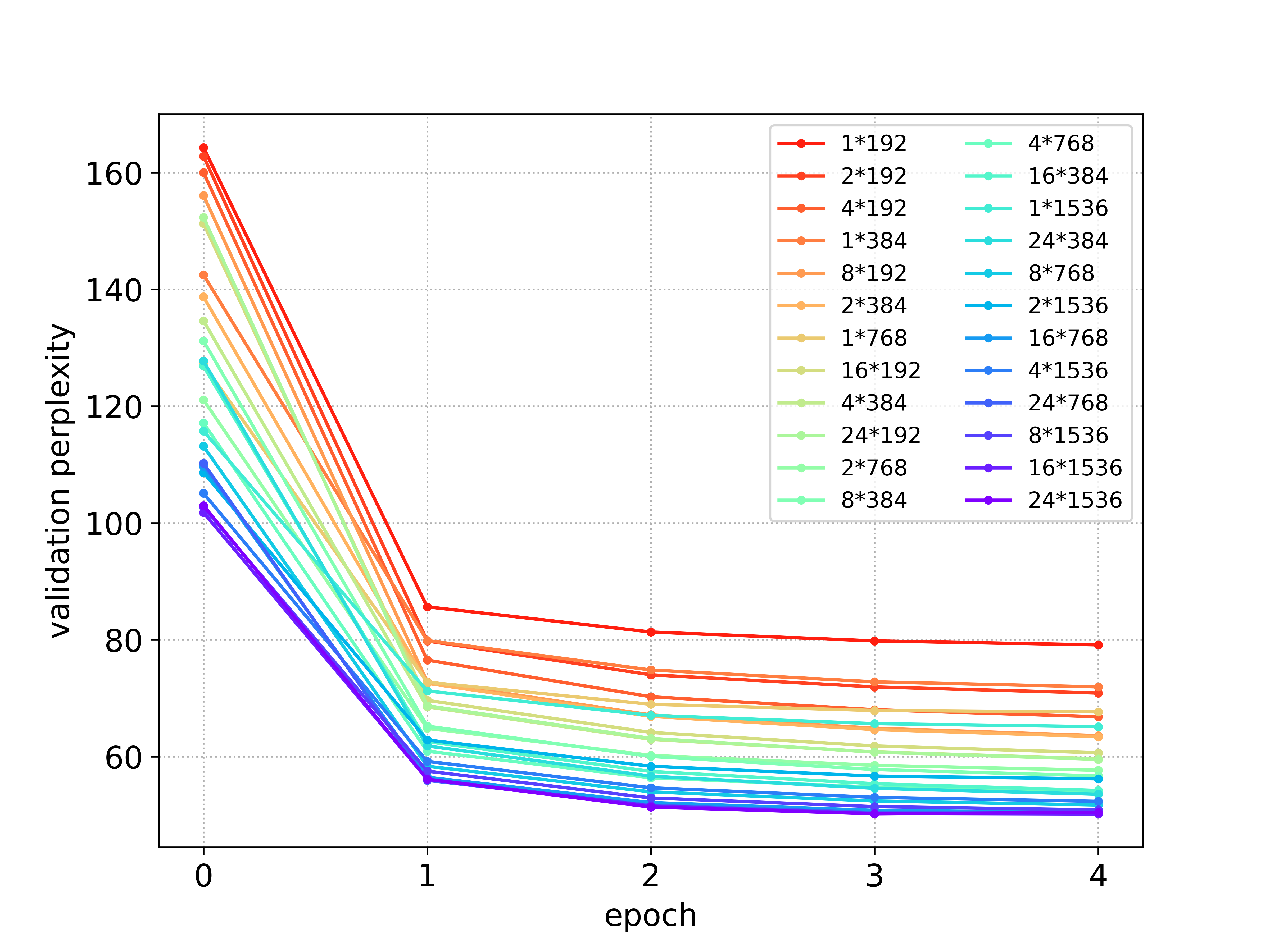}
    \caption{Validation perplexity by configuration and epoch on the development set of the BabyLM corpus.}
    \label{fig:pp_by_model}
\end{figure}

\section{Detailed resutls: Reading time experiments}
\label{sec:appendix_rts}
\def\arraystretch{1.5}%
\begin{table}[!h]
    \centering
        \begin{tabular}{l|c c c}
        \toprule
        \textbf{Corpus} & \textbf{Step} & \textbf{Spearman's} $\rho$ & \textbf{p-value} \\
        \midrule
        babylm & 500 & -0.5913 & 0.0097 \\
        babylm & 1000 & -0.6285 & 0.0052 \\
        babylm & 2000 & -0.7833 & 0.0001 \\
        babylm & 3000 & -0.7874 & 0.0001 \\
        babylm & 1 & -0.7915 & 0.0001 \\
        babylm & 5 & -0.614 & 0.0067 \\
        \midrule
        wikitext-103 & 500 & 0.0815 & 0.7478 \\
        wikitext-103 & 1000 & -0.4241 & 0.0794 \\
        wikitext-103 & 2000 & -0.7482 & 0.0004 \\
        wikitext-103 & 3000 & -0.7441 & 0.0004 \\
        wikitext-103 & 1 & -0.7172 & 0.0008 \\
        wikitext-103 & 5 & -0.7523 & 0.0003 \\
        \bottomrule
\end{tabular}

    \caption{Spearman's $\rho$ of model size (in terms of number of parameters) and $\Delta$ log-likelihood over the baseline LME model. Steps 1 and 5 refer to the first and fifth epoch.}
    \label{tab:detailed_rts_results}
\end{table}

\newpage
\section{Detailed results: BabyLM challenge tasks}

\begin{figure}[!h]
    \centering
    \includegraphics[width=\textwidth]{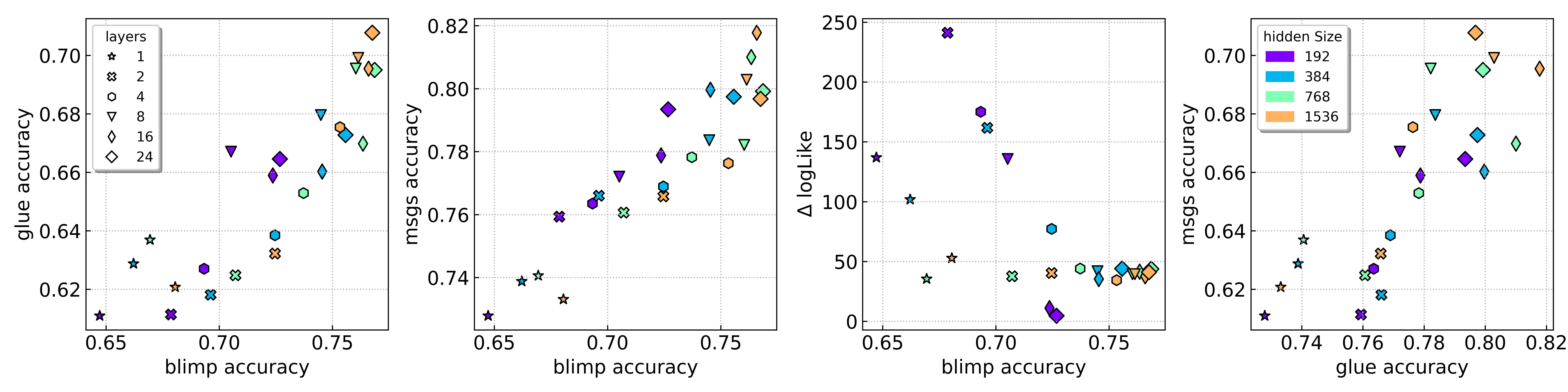}
    \caption{Correlation of LM performance on BLiMP vs. GLUE, BLiMP vs. MSGS, GLUE vs. MSGS.}
    \label{fig:blimp_vs_others}
\end{figure}

\fontsize{2pt}{2pt}\selectfont
\label{sec:appendix_b}
\begin{sidewaystable}
[!t]\label{tab:results_by_tasks_and_models}
    \footnotesize
        \caption{BLiMP accuracy by task and model}
        \begin{adjustbox}{width=0.9\textwidth, center}
        \begin{tabular}{l|c  c  c  c  c  c  c  c  c  c  c  c  c  c  c  c  c  c  c  c  c  c  c  c  c c}
\hline
\textbf{Task} & \textbf{1*192} & \textbf{1*384} & \textbf{1*768} & \textbf{1*1536} & \textbf{2*192} & \textbf{2*384} & \textbf{2*768} & \textbf{2*1536} & \textbf{4*192} & \textbf{4*384} & \textbf{4*768} & \textbf{4*1536} & \textbf{8*192} & \textbf{8*384} & \textbf{8*768} & \textbf{8*1536} & \textbf{16*192} & \textbf{16*384} & \textbf{16*768} & \textbf{16*1536} & \textbf{24*192} & \textbf{24*384} & \textbf{24*768} & \textbf{24*1536} & \textbf{OPT-125m (baseline)}\\
\hline\hline
anaphor-agreement & $0.95 \pm 0.01$ & $0.96 \pm 0.00$ & $0.94 \pm 0.01$ & $0.95 \pm 0.00$ & $0.94 \pm 0.01$ & $0.96 \pm 0.00$ & $0.97 \pm 0.00$ & $0.97 \pm 0.00$ & $0.96 \pm 0.00$ & $0.97 \pm 0.00$ & $0.97 \pm 0.00$ & $0.98 \pm 0.00$ & $0.95 \pm 0.01$ & $0.97 \pm 0.01$ & $0.99 \pm 0.00$ & $0.98 \pm 0.00$ & $0.96 \pm 0.01$ & $0.97 \pm 0.00$ & $0.98 \pm 0.00$ & $0.97 \pm 0.00$ & $0.96 \pm 0.01$ & $0.97 \pm 0.01$ & $0.98 \pm 0.00$ & $0.98 \pm 0.00$ & $94.9$ \\
argument-structure & $0.70 \pm 0.00$ & $0.71 \pm 0.00$ & $0.72 \pm 0.00$ & $0.72 \pm 0.00$ & $0.76 \pm 0.00$ & $0.76 \pm 0.00$ & $0.77 \pm 0.00$ & $0.77 \pm 0.00$ & $0.76 \pm 0.00$ & $0.78 \pm 0.00$ & $0.79 \pm 0.00$ & $0.79 \pm 0.00$ & $0.77 \pm 0.00$ & $0.79 \pm 0.00$ & $0.79 \pm 0.00$ & $0.79 \pm 0.00$ & $0.78 \pm 0.00$ & $0.80 \pm 0.00$ & $0.80 \pm 0.00$ & $0.80 \pm 0.00$ & $0.79 \pm 0.00$ & $0.80 \pm 0.00$ & $0.80 \pm 0.00$ & $0.79 \pm 0.01$ & $73.8$ \\
binding & $0.67 \pm 0.00$ & $0.67 \pm 0.00$ & $0.68 \pm 0.01$ & $0.68 \pm 0.00$ & $0.69 \pm 0.00$ & $0.69 \pm 0.02$ & $0.72 \pm 0.00$ & $0.71 \pm 0.02$ & $0.72 \pm 0.02$ & $0.74 \pm 0.01$ & $0.74 \pm 0.00$ & $0.72 \pm 0.00$ & $0.72 \pm 0.00$ & $0.74 \pm 0.00$ & $0.74 \pm 0.01$ & $0.75 \pm 0.00$ & $0.73 \pm 0.03$ & $0.72 \pm 0.00$ & $0.76 \pm 0.00$ & $0.76 \pm 0.01$ & $0.74 \pm 0.00$ & $0.74 \pm 0.00$ & $0.76 \pm 0.01$ & $0.77 \pm 0.01$ & $73.8$\\
control-raising & $0.64 \pm 0.01$ & $0.66 \pm 0.00$ & $0.68 \pm 0.01$ & $0.67 \pm 0.00$ & $0.67 \pm 0.00$ & $0.69 \pm 0.00$ & $0.72 \pm 0.00$ & $0.73 \pm 0.00$ & $0.69 \pm 0.00$ & $0.70 \pm 0.00$ & $0.73 \pm 0.00$ & $0.73 \pm 0.00$ & $0.73 \pm 0.00$ & $0.74 \pm 0.00$ & $0.74 \pm 0.01$ & $0.74 \pm 0.00$ & $0.73 \pm 0.00$ & $0.74 \pm 0.01$ & $0.72 \pm 0.00$ & $0.75 \pm 0.00$ & $0.72 \pm 0.00$ & $0.75 \pm 0.00$ & $0.74 \pm 0.00$ & $0.76 \pm 0.01$ & $72.2$\\
determiner-noun-agreement & $0.85 \pm 0.00$ & $0.85 \pm 0.01$ & $0.87 \pm 0.00$ & $0.88 \pm 0.00$ & $0.91 \pm 0.00$ & $0.91 \pm 0.00$ & $0.92 \pm 0.00$ & $0.93 \pm 0.00$ & $0.90 \pm 0.00$ & $0.93 \pm 0.00$ & $0.93 \pm 0.01$ & $0.94 \pm 0.00$ & $0.92 \pm 0.00$ & $0.95 \pm 0.00$ & $0.94 \pm 0.00$ & $0.95 \pm 0.00$ & $0.93 \pm 0.00$ & $0.94 \pm 0.00$ & $0.94 \pm 0.00$ & $0.95 \pm 0.00$ & $0.94 \pm 0.00$ & $0.95 \pm 0.00$ & $0.94 \pm 0.00$ & $0.94 \pm 0.00$ & $93.1$ \\
ellipsis & $0.57 \pm 0.01$ & $0.62 \pm 0.01$ & $0.63 \pm 0.00$ & $0.64 \pm 0.01$ & $0.62 \pm 0.02$ & $0.63 \pm 0.00$ & $0.70 \pm 0.00$ & $0.72 \pm 0.01$ & $0.66 \pm 0.01$ & $0.69 \pm 0.01$ & $0.75 \pm 0.01$ & $0.78 \pm 0.02$ & $0.71 \pm 0.00$ & $0.76 \pm 0.01$ & $0.82 \pm 0.01$ & $0.82 \pm 0.01$ & $0.75 \pm 0.01$ & $0.80 \pm 0.00$ & $0.82 \pm 0.01$ & $0.82 \pm 0.03$ & $0.76 \pm 0.01$ & $0.81 \pm 0.00$ & $0.84 \pm 0.02$ & $0.82 \pm 0.01$ & $80.5$\\
filler-gap & $0.64 \pm 0.01$ & $0.65 \pm 0.00$ & $0.65 \pm 0.00$ & $0.65 \pm 0.01$ & $0.67 \pm 0.00$ & $0.68 \pm 0.00$ & $0.70 \pm 0.01$ & $0.72 \pm 0.00$ & $0.68 \pm 0.00$ & $0.71 \pm 0.00$ & $0.73 \pm 0.00$ & $0.75 \pm 0.00$ & $0.70 \pm 0.01$ & $0.72 \pm 0.00$ & $0.74 \pm 0.00$ & $0.74 \pm 0.00$ & $0.72 \pm 0.00$ & $0.74 \pm 0.01$ & $0.75 \pm 0.00$ & $0.75 \pm 0.01$ & $0.73 \pm 0.00$ & $0.74 \pm 0.00$ & $0.76 \pm 0.00$ & $0.75 \pm 0.01$ & $73.6$\\
irregular-forms & $0.88 \pm 0.01$ & $0.90 \pm 0.01$ & $0.89 \pm 0.01$ & $0.89 \pm 0.01$ & $0.93 \pm 0.00$ & $0.93 \pm 0.00$ & $0.92 \pm 0.01$ & $0.93 \pm 0.00$ & $0.92 \pm 0.00$ & $0.91 \pm 0.01$ & $0.91 \pm 0.00$ & $0.90 \pm 0.00$ & $0.91 \pm 0.02$ & $0.90 \pm 0.01$ & $0.92 \pm 0.01$ & $0.92 \pm 0.01$ & $0.93 \pm 0.00$ & $0.92 \pm 0.02$ & $0.92 \pm 0.01$ & $0.92 \pm 0.00$ & $0.89 \pm 0.03$ & $0.90 \pm 0.02$ & $0.92 \pm 0.00$ & $0.87 \pm 0.02$ & $80.8$ \\
island-effects & $0.54 \pm 0.01$ & $0.55 \pm 0.01$ & $0.56 \pm 0.01$ & $0.61 \pm 0.02$ & $0.50 \pm 0.03$ & $0.55 \pm 0.01$ & $0.52 \pm 0.01$ & $0.57 \pm 0.02$ & $0.56 \pm 0.03$ & $0.60 \pm 0.01$ & $0.63 \pm 0.00$ & $0.65 \pm 0.01$ & $0.55 \pm 0.03$ & $0.64 \pm 0.02$ & $0.66 \pm 0.02$ & $0.67 \pm 0.00$ & $0.61 \pm 0.02$ & $0.65 \pm 0.01$ & $0.65 \pm 0.00$ & $0.69 \pm 0.01$ & $0.62 \pm 0.02$ & $0.65 \pm 0.01$ & $0.68 \pm 0.00$ & $0.70 \pm 0.00$ & $51.6$ \\
npi-licensing & $0.53 \pm 0.01$ & $0.52 \pm 0.00$ & $0.52 \pm 0.00$ & $0.56 \pm 0.00$ & $0.57 \pm 0.00$ & $0.64 \pm 0.01$ & $0.60 \pm 0.02$ & $0.63 \pm 0.01$ & $0.58 \pm 0.04$ & $0.65 \pm 0.00$ & $0.67 \pm 0.01$ & $0.72 \pm 0.00$ & $0.63 \pm 0.00$ & $0.65 \pm 0.01$ & $0.68 \pm 0.00$ & $0.71 \pm 0.00$ & $0.63 \pm 0.01$ & $0.69 \pm 0.03$ & $0.72 \pm 0.01$ & $0.70 \pm 0.02$ & $0.66 \pm 0.02$ & $0.75 \pm 0.00$ & $0.70 \pm 0.01$ & $0.70 \pm 0.01$ & $51.6$\\
quantifiers & $0.71 \pm 0.01$ & $0.71 \pm 0.01$ & $0.74 \pm 0.01$ & $0.73 \pm 0.02$ & $0.75 \pm 0.01$ & $0.71 \pm 0.01$ & $0.69 \pm 0.01$ & $0.71 \pm 0.02$ & $0.73 \pm 0.01$ & $0.75 \pm 0.00$ & $0.71 \pm 0.03$ & $0.74 \pm 0.02$ & $0.72 \pm 0.02$ & $0.76 \pm 0.01$ & $0.73 \pm 0.01$ & $0.70 \pm 0.00$ & $0.69 \pm 0.02$ & $0.71 \pm 0.02$ & $0.74 \pm 0.00$ & $0.74 \pm 0.00$ & $0.66 \pm 0.00$ & $0.69 \pm 0.01$ & $0.77 \pm 0.01$ & $0.75 \pm 0.00$ & $74.5$ \\
subject-verb-agreement & $0.62 \pm 0.00$ & $0.64 \pm 0.00$ & $0.65 \pm 0.00$ & $0.67 \pm 0.00$ & $0.71 \pm 0.00$ & $0.76 \pm 0.00$ & $0.79 \pm 0.01$ & $0.81 \pm 0.01$ & $0.75 \pm 0.00$ & $0.83 \pm 0.00$ & $0.85 \pm 0.00$ & $0.87 \pm 0.00$ & $0.81 \pm 0.00$ & $0.86 \pm 0.00$ & $0.87 \pm 0.00$ & $0.88 \pm 0.01$ & $0.84 \pm 0.01$ & $0.88 \pm 0.00$ & $0.87 \pm 0.00$ & $0.88 \pm 0.00$ & $0.84 \pm 0.00$ & $0.87 \pm 0.01$ & $0.87 \pm 0.00$ & $0.88 \pm 0.01$ & $77.3$ \\
hypernym & $0.51 \pm 0.00$ & $0.50 \pm 0.01$ & $0.49 \pm 0.00$ & $0.48 \pm 0.00$ & $0.49 \pm 0.00$ & $0.48 \pm 0.00$ & $0.47 \pm 0.01$ & $0.48 \pm 0.00$ & $0.49 \pm 0.01$ & $0.48 \pm 0.00$ & $0.50 \pm 0.00$ & $0.48 \pm 0.01$ & $0.48 \pm 0.00$ & $0.48 \pm 0.01$ & $0.48 \pm 0.00$ & $0.48 \pm 0.00$ & $0.49 \pm 0.01$ & $0.48 \pm 0.00$ & $0.46 \pm 0.00$ & $0.46 \pm 0.00$ & $0.48 \pm 0.00$ & $0.48 \pm 0.00$ & $0.47 \pm 0.01$ & $0.46 \pm 0.00$ & $46.3$\\
qa-congruence-easy & $0.47 \pm 0.03$ & $0.55 \pm 0.02$ & $0.52 \pm 0.01$ & $0.59 \pm 0.00$ & $0.50 \pm 0.02$ & $0.58 \pm 0.02$ & $0.60 \pm 0.01$ & $0.66 \pm 0.00$ & $0.58 \pm 0.01$ & $0.64 \pm 0.02$ & $0.67 \pm 0.04$ & $0.71 \pm 0.01$ & $0.59 \pm 0.03$ & $0.66 \pm 0.01$ & $0.69 \pm 0.01$ & $0.68 \pm 0.03$ & $0.59 \pm 0.02$ & $0.66 \pm 0.02$ & $0.71 \pm 0.01$ & $0.70 \pm 0.04$ & $0.62 \pm 0.01$ & $0.69 \pm 0.03$ & $0.72 \pm 0.00$ & $0.68 \pm 0.03$ & $76.5$ \\
qa-congruence-tricky & $0.29 \pm 0.00$ & $0.36 \pm 0.01$ & $0.35 \pm 0.05$ & $0.42 \pm 0.02$ & $0.35 \pm 0.01$ & $0.39 \pm 0.02$ & $0.42 \pm 0.01$ & $0.41 \pm 0.01$ & $0.33 \pm 0.04$ & $0.41 \pm 0.04$ & $0.45 \pm 0.01$ & $0.46 \pm 0.00$ & $0.32 \pm 0.00$ & $0.48 \pm 0.02$ & $0.52 \pm 0.01$ & $0.48 \pm 0.01$ & $0.39 \pm 0.03$ & $0.40 \pm 0.01$ & $0.50 \pm 0.00$ & $0.51 \pm 0.00$ & $0.38 \pm 0.03$ & $0.50 \pm 0.04$ & $0.46 \pm 0.00$ & $0.53 \pm 0.04$ & $47.9$ \\
subject-aux-inversion & $0.83 \pm 0.02$ & $0.81 \pm 0.00$ & $0.83 \pm 0.01$ & $0.81 \pm 0.00$ & $0.83 \pm 0.02$ & $0.82 \pm 0.01$ & $0.80 \pm 0.00$ & $0.86 \pm 0.01$ & $0.84 \pm 0.00$ & $0.82 \pm 0.02$ & $0.80 \pm 0.02$ & $0.82 \pm 0.02$ & $0.81 \pm 0.01$ & $0.83 \pm 0.01$ & $0.84 \pm 0.01$ & $0.86 \pm 0.01$ & $0.85 \pm 0.02$ & $0.83 \pm 0.01$ & $0.84 \pm 0.00$ & $0.85 \pm 0.02$ & $0.86 \pm 0.01$ & $0.83 \pm 0.02$ & $0.87 \pm 0.02$ & $0.88 \pm 0.01$ & $85.3$\\
turn-taking & $0.61 \pm 0.02$ & $0.60 \pm 0.01$ & $0.65 \pm 0.01$ & $0.61 \pm 0.01$ & $0.66 \pm 0.01$ & $0.65 \pm 0.02$ & $0.71 \pm 0.02$ & $0.71 \pm 0.00$ & $0.64 \pm 0.01$ & $0.71 \pm 0.01$ & $0.71 \pm 0.01$ & $0.75 \pm 0.02$ & $0.66 \pm 0.02$ & $0.73 \pm 0.01$ & $0.78 \pm 0.01$ & $0.79 \pm 0.01$ & $0.69 \pm 0.01$ & $0.74 \pm 0.01$ & $0.80 \pm 0.01$ & $0.77 \pm 0.01$ & $0.72 \pm 0.01$ & $0.73 \pm 0.00$ & $0.78 \pm 0.01$ & $0.78 \pm 0.00$ & $82.9$ \\
\hline
\textbf{Average} & $0.65 \pm 0.03$ & $0.66 \pm 0.03$ & $0.67 \pm 0.03$ & $0.68 \pm 0.02$ & $0.68 \pm 0.03$ & $0.7 \pm 0.03$ & $0.71 \pm 0.03$ & $0.72 \pm 0.03$ & $0.69 \pm 0.03$ & $0.72 \pm 0.02$ & $0.74 \pm 0.02$ & $0.75 \pm 0.02$ & $0.71 \pm 0.03$ & $0.74 \pm 0.02$ & $0.76 \pm 0.02$ & $0.76 \pm 0.02$ & $0.72 \pm 0.03$ & $0.75 \pm 0.03$ & $0.76 \pm 0.02$ & $0.77 \pm 0.02$ & $0.73 \pm 0.03$ & $0.76 \pm 0.02$ & $0.77 \pm 0.02$ & $0.77 \pm 0.02$ & $73.11$ \\
\hline
\end{tabular}
        \end{adjustbox}
        \vspace{2\baselineskip}
        \caption{GLUE accuracy by task and model}
        \begin{adjustbox}{width=0.9\textwidth, center}
        \begin{tabular}{l|c  c  c  c  c  c  c  c  c  c  c  c  c  c  c  c  c  c  c  c  c  c  c  c  c c }
\hline
\textbf{Task} & \textbf{1*192} & \textbf{1*384} & \textbf{1*768} & \textbf{1*1536} & \textbf{2*192} & \textbf{2*384} & \textbf{2*768} & \textbf{2*1536} & \textbf{4*192} & \textbf{4*384} & \textbf{4*768} & \textbf{4*1536} & \textbf{8*192} & \textbf{8*384} & \textbf{8*768} & \textbf{8*1536} & \textbf{16*192} & \textbf{16*384} & \textbf{16*768} & \textbf{16*1536} & \textbf{24*192} & \textbf{24*384} & \textbf{24*768} & \textbf{24*1536} & \textbf{OPT-125m (baseline)} \\
\hline\hline
boolq & $0.59 \pm 0.00$ & $0.60 \pm 0.00$ & $0.59 \pm 0.00$ & $0.60 \pm 0.00$ & $0.58 \pm 0.00$ & $0.60 \pm 0.00$ & $0.59 \pm 0.00$ & $0.60 \pm 0.00$ & $0.59 \pm 0.00$ & $0.60 \pm 0.00$ & $0.60 \pm 0.01$ & $0.59 \pm 0.00$ & $0.60 \pm 0.00$ & $0.60 \pm 0.01$ & $0.60 \pm 0.00$ & $0.61 \pm 0.00$ & $0.60 \pm 0.00$ & $0.59 \pm 0.00$ & $0.59 \pm 0.01$ & $0.61 \pm 0.01$ & $0.60 \pm 0.00$ & $0.60 \pm 0.00$ & $0.60 \pm 0.00$ & $0.64 \pm 0.01$ & $66.0$ \\
cola & $0.69 \pm 0.00$ & $0.70 \pm 0.00$ & $0.69 \pm 0.00$ & $0.70 \pm 0.00$ & $0.69 \pm 0.00$ & $0.69 \pm 0.00$ & $0.69 \pm 0.00$ & $0.69 \pm 0.00$ & $0.69 \pm 0.00$ & $0.69 \pm 0.00$ & $0.70 \pm 0.00$ & $0.70 \pm 0.00$ & $0.69 \pm 0.00$ & $0.69 \pm 0.00$ & $0.70 \pm 0.00$ & $0.74 \pm 0.00$ & $0.70 \pm 0.00$ & $0.70 \pm 0.00$ & $0.71 \pm 0.00$ & $0.74 \pm 0.00$ & $0.70 \pm 0.00$ & $0.71 \pm 0.00$ & $0.75 \pm 0.00$ & $0.76 \pm 0.00$ & $36$ \\
mnli & $0.55 \pm 0.00$ & $0.56 \pm 0.00$ & $0.58 \pm 0.00$ & $0.53 \pm 0.00$ & $0.52 \pm 0.01$ & $0.54 \pm 0.00$ & $0.55 \pm 0.00$ & $0.57 \pm 0.00$ & $0.54 \pm 0.01$ & $0.59 \pm 0.02$ & $0.64 \pm 0.00$ & $0.65 \pm 0.00$ & $0.67 \pm 0.00$ & $0.71 \pm 0.00$ & $0.72 \pm 0.00$ & $0.74 \pm 0.00$ & $0.63 \pm 0.00$ & $0.66 \pm 0.00$ & $0.67 \pm 0.00$ & $0.74 \pm 0.00$ & $0.64 \pm 0.00$ & $0.66 \pm 0.00$ & $0.74 \pm 0.00$ & $0.75 \pm 0.00$ & $70.1$\\
mnli-mm & $0.56 \pm 0.00$ & $0.57 \pm 0.00$ & $0.57 \pm 0.00$ & $0.54 \pm 0.00$ & $0.53 \pm 0.00$ & $0.54 \pm 0.00$ & $0.55 \pm 0.00$ & $0.58 \pm 0.01$ & $0.56 \pm 0.01$ & $0.61 \pm 0.02$ & $0.67 \pm 0.00$ & $0.68 \pm 0.00$ & $0.69 \pm 0.00$ & $0.72 \pm 0.00$ & $0.73 \pm 0.00$ & $0.74 \pm 0.00$ & $0.65 \pm 0.00$ & $0.68 \pm 0.00$ & $0.70 \pm 0.00$ & $0.76 \pm 0.00$ & $0.67 \pm 0.00$ & $0.69 \pm 0.00$ & $0.75 \pm 0.00$ & $0.76 \pm 0.00$ & $71.9$\\
mrpc & $0.69 \pm 0.00$ & $0.69 \pm 0.00$ & $0.69 \pm 0.00$ & $0.69 \pm 0.00$ & $0.69 \pm 0.00$ & $0.69 \pm 0.00$ & $0.69 \pm 0.00$ & $0.69 \pm 0.00$ & $0.69 \pm 0.00$ & $0.69 \pm 0.00$ & $0.69 \pm 0.00$ & $0.69 \pm 0.00$ & $0.69 \pm 0.00$ & $0.69 \pm 0.00$ & $0.69 \pm 0.00$ & $0.69 \pm 0.00$ & $0.70 \pm 0.00$ & $0.69 \pm 0.00$ & $0.69 \pm 0.00$ & $0.70 \pm 0.01$ & $0.69 \pm 0.00$ & $0.69 \pm 0.00$ & $0.69 \pm 0.00$ & $0.68 \pm 0.01$ & $82.1$\\
multirc & $0.50 \pm 0.00$ & $0.52 \pm 0.00$ & $0.51 \pm 0.00$ & $0.51 \pm 0.01$ & $0.50 \pm 0.01$ & $0.49 \pm 0.02$ & $0.51 \pm 0.01$ & $0.52 \pm 0.01$ & $0.51 \pm 0.01$ & $0.51 \pm 0.02$ & $0.52 \pm 0.01$ & $0.53 \pm 0.00$ & $0.51 \pm 0.01$ & $0.50 \pm 0.01$ & $0.53 \pm 0.01$ & $0.51 \pm 0.00$ & $0.52 \pm 0.01$ & $0.48 \pm 0.01$ & $0.50 \pm 0.01$ & $0.52 \pm 0.02$ & $0.50 \pm 0.00$ & $0.50 \pm 0.01$ & $0.50 \pm 0.00$ & $0.51 \pm 0.01$ & $61.1$\\
qnli & $0.60 \pm 0.00$ & $0.61 \pm 0.00$ & $0.60 \pm 0.00$ & $0.60 \pm 0.00$ & $0.60 \pm 0.00$ & $0.61 \pm 0.00$ & $0.61 \pm 0.00$ & $0.62 \pm 0.00$ & $0.62 \pm 0.00$ & $0.69 \pm 0.05$ & $0.78 \pm 0.00$ & $0.80 \pm 0.00$ & $0.74 \pm 0.02$ & $0.79 \pm 0.00$ & $0.81 \pm 0.00$ & $0.82 \pm 0.01$ & $0.78 \pm 0.00$ & $0.80 \pm 0.00$ & $0.81 \pm 0.00$ & $0.82 \pm 0.00$ & $0.78 \pm 0.00$ & $0.80 \pm 0.00$ & $0.82 \pm 0.00$ & $0.83 \pm 0.00$ & $80.1$ \\
qqp & $0.76 \pm 0.00$ & $0.76 \pm 0.00$ & $0.77 \pm 0.00$ & $0.74 \pm 0.00$ & $0.71 \pm 0.00$ & $0.73 \pm 0.00$ & $0.74 \pm 0.00$ & $0.75 \pm 0.00$ & $0.73 \pm 0.01$ & $0.75 \pm 0.01$ & $0.78 \pm 0.00$ & $0.80 \pm 0.00$ & $0.81 \pm 0.00$ & $0.83 \pm 0.00$ & $0.84 \pm 0.00$ & $0.85 \pm 0.00$ & $0.77 \pm 0.00$ & $0.79 \pm 0.00$ & $0.80 \pm 0.00$ & $0.85 \pm 0.00$ & $0.78 \pm 0.00$ & $0.79 \pm 0.00$ & $0.84 \pm 0.00$ & $0.86 \pm 0.00$ & $77.8$\\
rte & $0.47 \pm 0.00$ & $0.57 \pm 0.00$ & $0.61 \pm 0.00$ & $0.56 \pm 0.04$ & $0.53 \pm 0.01$ & $0.50 \pm 0.01$ & $0.55 \pm 0.01$ & $0.52 \pm 0.02$ & $0.49 \pm 0.00$ & $0.51 \pm 0.02$ & $0.50 \pm 0.00$ & $0.58 \pm 0.04$ & $0.56 \pm 0.01$ & $0.53 \pm 0.04$ & $0.56 \pm 0.02$ & $0.51 \pm 0.00$ & $0.51 \pm 0.00$ & $0.53 \pm 0.00$ & $0.53 \pm 0.02$ & $0.52 \pm 0.01$ & $0.52 \pm 0.01$ & $0.56 \pm 0.01$ & $0.56 \pm 0.04$ & $0.54 \pm 0.01$ & $67.7$ \\
sst2 & $0.82 \pm 0.00$ & $0.83 \pm 0.00$ & $0.85 \pm 0.00$ & $0.84 \pm 0.01$ & $0.85 \pm 0.00$ & $0.85 \pm 0.00$ & $0.86 \pm 0.00$ & $0.86 \pm 0.00$ & $0.86 \pm 0.00$ & $0.85 \pm 0.00$ & $0.86 \pm 0.00$ & $0.86 \pm 0.01$ & $0.86 \pm 0.00$ & $0.86 \pm 0.00$ & $0.88 \pm 0.00$ & $0.87 \pm 0.00$ & $0.84 \pm 0.00$ & $0.85 \pm 0.00$ & $0.87 \pm 0.00$ & $0.89 \pm 0.00$ & $0.85 \pm 0.00$ & $0.87 \pm 0.00$ & $0.88 \pm 0.00$ & $0.89 \pm 0.00$ & $86.6$ \\
wsc & $0.48 \pm 0.00$ & $0.52 \pm 0.00$ & $0.54 \pm 0.00$ & $0.53 \pm 0.03$ & $0.52 \pm 0.03$ & $0.55 \pm 0.03$ & $0.53 \pm 0.01$ & $0.55 \pm 0.03$ & $0.61 \pm 0.01$ & $0.52 \pm 0.00$ & $0.46 \pm 0.00$ & $0.55 \pm 0.03$ & $0.51 \pm 0.04$ & $0.56 \pm 0.00$ & $0.59 \pm 0.02$ & $0.61 \pm 0.00$ & $0.57 \pm 0.01$ & $0.49 \pm 0.01$ & $0.49 \pm 0.01$ & $0.50 \pm 0.02$ & $0.58 \pm 0.00$ & $0.52 \pm 0.02$ & $0.51 \pm 0.02$ & $0.57 \pm 0.00$ & $59.0$ \\
\hline
\textbf{Average} & $0.61 \pm 0.03$ & $0.63 \pm 0.03$ & $0.64 \pm 0.03$ & $0.62 \pm 0.02$ & $0.61 \pm 0.02$ & $0.62 \pm 0.02$ & $0.62 \pm 0.02$ & $0.63 \pm 0.02$ & $0.63 \pm 0.02$ & $0.64 \pm 0.02$ & $0.65 \pm 0.03$ & $0.68 \pm 0.02$ & $0.67 \pm 0.02$ & $0.68 \pm 0.02$ & $0.7 \pm 0.02$ & $0.7 \pm 0.03$ & $0.66 \pm 0.02$ & $0.66 \pm 0.03$ & $0.67 \pm 0.03$ & $0.7 \pm 0.03$ & $0.66 \pm 0.02$ & $0.67 \pm 0.02$ & $0.7 \pm 0.03$ & $0.71 \pm 0.03$ & $68.95$ \\
\hline
\end{tabular}
        \end{adjustbox}
        \vspace{2\baselineskip}
        \caption{MSGS accuracy by task and model}
                \begin{adjustbox}{width=0.9\textwidth, center}
        \begin{tabular}{l|c  c  c  c  c  c  c  c  c  c  c  c  c  c  c  c  c  c  c  c  c  c  c  c  c }
\hline
\textbf{Task} & \textbf{1*192} & \textbf{1*384} & \textbf{1*768} & \textbf{1*1536} & \textbf{2*192} & \textbf{2*384} & \textbf{2*768} & \textbf{2*1536} & \textbf{4*192} & \textbf{4*384} & \textbf{4*768} & \textbf{4*1536} & \textbf{8*192} & \textbf{8*384} & \textbf{8*768} & \textbf{8*1536} & \textbf{16*192} & \textbf{16*384} & \textbf{16*768} & \textbf{16*1536} & \textbf{24*192} & \textbf{24*384} & \textbf{24*768} & \textbf{24*1536} \\
\hline\hline
main-verb-control & $0.63 \pm 0.01$ & $0.66 \pm 0.01$ & $0.65 \pm 0.00$ & $0.66 \pm 0.00$ & $0.78 \pm 0.04$ & $0.90 \pm 0.01$ & $0.91 \pm 0.00$ & $0.93 \pm 0.01$ & $0.92 \pm 0.00$ & $0.95 \pm 0.01$ & $0.99 \pm 0.00$ & $0.99 \pm 0.00$ & $0.99 \pm 0.00$ & $1.00 \pm 0.00$ & $1.00 \pm 0.00$ & $1.00 \pm 0.00$ & $0.99 \pm 0.00$ & $0.99 \pm 0.00$ & $1.00 \pm 0.00$ & $1.00 \pm 0.00$ & $0.99 \pm 0.00$ & $1.00 \pm 0.00$ & $1.00 \pm 0.00$ & $1.00 \pm 0.00$ \\
control-raising-control & $0.82 \pm 0.01$ & $0.85 \pm 0.00$ & $0.84 \pm 0.02$ & $0.90 \pm 0.00$ & $0.88 \pm 0.01$ & $0.91 \pm 0.01$ & $0.93 \pm 0.00$ & $0.93 \pm 0.00$ & $0.92 \pm 0.00$ & $0.94 \pm 0.01$ & $0.94 \pm 0.01$ & $0.96 \pm 0.01$ & $0.92 \pm 0.00$ & $0.93 \pm 0.01$ & $0.96 \pm 0.01$ & $0.96 \pm 0.00$ & $0.93 \pm 0.01$ & $0.95 \pm 0.01$ & $0.97 \pm 0.00$ & $0.97 \pm 0.00$ & $0.94 \pm 0.01$ & $0.95 \pm 0.01$ & $0.96 \pm 0.01$ & $0.96 \pm 0.00$ \\
syntactic-category-control & $0.82 \pm 0.02$ & $0.82 \pm 0.00$ & $0.84 \pm 0.00$ & $0.79 \pm 0.01$ & $0.90 \pm 0.01$ & $0.88 \pm 0.01$ & $0.82 \pm 0.02$ & $0.83 \pm 0.01$ & $0.81 \pm 0.01$ & $0.78 \pm 0.01$ & $0.83 \pm 0.01$ & $0.80 \pm 0.01$ & $0.83 \pm 0.01$ & $0.86 \pm 0.00$ & $0.83 \pm 0.00$ & $0.84 \pm 0.01$ & $0.83 \pm 0.00$ & $0.84 \pm 0.01$ & $0.87 \pm 0.02$ & $0.78 \pm 0.00$ & $0.83 \pm 0.01$ & $0.84 \pm 0.02$ & $0.83 \pm 0.00$ & $0.82 \pm 0.01$ \\
lexical-content-the-control & $0.79 \pm 0.01$ & $0.79 \pm 0.00$ & $0.80 \pm 0.02$ & $0.76 \pm 0.00$ & $0.80 \pm 0.02$ & $0.76 \pm 0.00$ & $0.75 \pm 0.00$ & $0.75 \pm 0.00$ & $0.81 \pm 0.01$ & $0.83 \pm 0.00$ & $0.79 \pm 0.01$ & $0.76 \pm 0.00$ & $0.79 \pm 0.01$ & $0.80 \pm 0.03$ & $0.77 \pm 0.01$ & $0.76 \pm 0.00$ & $0.78 \pm 0.01$ & $0.77 \pm 0.01$ & $0.76 \pm 0.00$ & $0.76 \pm 0.00$ & $0.88 \pm 0.04$ & $0.78 \pm 0.00$ & $0.76 \pm 0.01$ & $0.75 \pm 0.00$ \\
relative-position-control & $0.96 \pm 0.01$ & $0.98 \pm 0.00$ & $0.95 \pm 0.01$ & $0.92 \pm 0.00$ & $0.97 \pm 0.00$ & $0.97 \pm 0.00$ & $0.96 \pm 0.00$ & $0.97 \pm 0.00$ & $0.98 \pm 0.00$ & $0.98 \pm 0.00$ & $0.97 \pm 0.01$ & $0.99 \pm 0.00$ & $0.99 \pm 0.00$ & $1.00 \pm 0.00$ & $0.99 \pm 0.00$ & $0.99 \pm 0.00$ & $0.98 \pm 0.01$ & $1.00 \pm 0.00$ & $1.00 \pm 0.00$ & $0.99 \pm 0.00$ & $0.99 \pm 0.00$ & $0.98 \pm 0.01$ & $1.00 \pm 0.00$ & $0.99 \pm 0.00$ \\
main-verb-lexical-content-the & $0.67 \pm 0.00$ & $0.66 \pm 0.00$ & $0.66 \pm 0.00$ & $0.67 \pm 0.00$ & $0.67 \pm 0.00$ & $0.67 \pm 0.00$ & $0.67 \pm 0.00$ & $0.67 \pm 0.00$ & $0.68 \pm 0.00$ & $0.67 \pm 0.00$ & $0.68 \pm 0.01$ & $0.69 \pm 0.01$ & $0.67 \pm 0.00$ & $0.69 \pm 0.00$ & $0.68 \pm 0.01$ & $0.73 \pm 0.02$ & $0.69 \pm 0.01$ & $0.75 \pm 0.00$ & $0.71 \pm 0.00$ & $0.81 \pm 0.01$ & $0.71 \pm 0.01$ & $0.73 \pm 0.01$ & $0.77 \pm 0.02$ & $0.76 \pm 0.01$ \\
main-verb-relative-token-position & $0.66 \pm 0.00$ & $0.66 \pm 0.00$ & $0.66 \pm 0.00$ & $0.67 \pm 0.00$ & $0.67 \pm 0.00$ & $0.66 \pm 0.00$ & $0.67 \pm 0.00$ & $0.67 \pm 0.00$ & $0.67 \pm 0.00$ & $0.67 \pm 0.00$ & $0.69 \pm 0.01$ & $0.68 \pm 0.00$ & $0.66 \pm 0.00$ & $0.68 \pm 0.01$ & $0.70 \pm 0.01$ & $0.75 \pm 0.01$ & $0.69 \pm 0.01$ & $0.76 \pm 0.01$ & $0.74 \pm 0.01$ & $0.78 \pm 0.01$ & $0.72 \pm 0.02$ & $0.74 \pm 0.01$ & $0.77 \pm 0.00$ & $0.78 \pm 0.02$ \\
syntactic-category-lexical-content-the & $0.67 \pm 0.03$ & $0.70 \pm 0.00$ & $0.73 \pm 0.01$ & $0.71 \pm 0.00$ & $0.70 \pm 0.01$ & $0.67 \pm 0.03$ & $0.67 \pm 0.03$ & $0.70 \pm 0.07$ & $0.62 \pm 0.02$ & $0.65 \pm 0.00$ & $0.68 \pm 0.01$ & $0.64 \pm 0.01$ & $0.66 \pm 0.02$ & $0.65 \pm 0.02$ & $0.65 \pm 0.02$ & $0.68 \pm 0.02$ & $0.67 \pm 0.02$ & $0.67 \pm 0.03$ & $0.68 \pm 0.01$ & $0.63 \pm 0.00$ & $0.64 \pm 0.01$ & $0.67 \pm 0.01$ & $0.66 \pm 0.00$ & $0.66 \pm 0.01$ \\
syntactic-category-relative-position & $0.66 \pm 0.00$ & $0.66 \pm 0.00$ & $0.68 \pm 0.00$ & $0.66 \pm 0.00$ & $0.65 \pm 0.01$ & $0.67 \pm 0.01$ & $0.66 \pm 0.00$ & $0.63 \pm 0.00$ & $0.65 \pm 0.01$ & $0.63 \pm 0.00$ & $0.65 \pm 0.00$ & $0.64 \pm 0.00$ & $0.64 \pm 0.01$ & $0.64 \pm 0.00$ & $0.63 \pm 0.00$ & $0.64 \pm 0.01$ & $0.63 \pm 0.01$ & $0.62 \pm 0.00$ & $0.63 \pm 0.00$ & $0.63 \pm 0.00$ & $0.62 \pm 0.00$ & $0.62 \pm 0.00$ & $0.62 \pm 0.01$ & $0.61 \pm 0.00$ \\
control-raising-lexical-content-the & $0.66 \pm 0.00$ & $0.67 \pm 0.00$ & $0.67 \pm 0.00$ & $0.66 \pm 0.00$ & $0.67 \pm 0.00$ & $0.67 \pm 0.00$ & $0.66 \pm 0.00$ & $0.67 \pm 0.00$ & $0.67 \pm 0.00$ & $0.67 \pm 0.00$ & $0.68 \pm 0.01$ & $0.70 \pm 0.01$ & $0.68 \pm 0.00$ & $0.68 \pm 0.00$ & $0.70 \pm 0.02$ & $0.79 \pm 0.01$ & $0.68 \pm 0.00$ & $0.75 \pm 0.00$ & $0.82 \pm 0.05$ & $0.91 \pm 0.01$ & $0.72 \pm 0.00$ & $0.75 \pm 0.03$ & $0.71 \pm 0.02$ & $0.82 \pm 0.01$ \\
control-raising-relative-token-position & $0.67 \pm 0.00$ & $0.67 \pm 0.00$ & $0.67 \pm 0.00$ & $0.68 \pm 0.00$ & $0.67 \pm 0.00$ & $0.67 \pm 0.00$ & $0.67 \pm 0.00$ & $0.67 \pm 0.00$ & $0.67 \pm 0.00$ & $0.67 \pm 0.00$ & $0.69 \pm 0.01$ & $0.69 \pm 0.00$ & $0.67 \pm 0.00$ & $0.69 \pm 0.01$ & $0.69 \pm 0.00$ & $0.69 \pm 0.00$ & $0.68 \pm 0.01$ & $0.69 \pm 0.00$ & $0.73 \pm 0.01$ & $0.74 \pm 0.01$ & $0.69 \pm 0.01$ & $0.71 \pm 0.01$ & $0.72 \pm 0.00$ & $0.71 \pm 0.00$ \\
\hline
\textbf{Average} & $0.73 \pm 0.02$ & $0.74 \pm 0.02$ & $0.74 \pm 0.02$ & $0.73 \pm 0.02$ & $0.76 \pm 0.02$ & $0.77 \pm 0.03$ & $0.76 \pm 0.02$ & $0.77 \pm 0.03$ & $0.76 \pm 0.03$ & $0.77 \pm 0.03$ & $0.78 \pm 0.03$ & $0.78 \pm 0.03$ & $0.77 \pm 0.03$ & $0.78 \pm 0.03$ & $0.78 \pm 0.03$ & $0.8 \pm 0.03$ & $0.78 \pm 0.03$ & $0.8 \pm 0.03$ & $0.81 \pm 0.03$ & $0.82 \pm 0.03$ & $0.79 \pm 0.03$ & $0.8 \pm 0.03$ & $0.8 \pm 0.03$ & $0.8 \pm 0.03$ \\
\hline
\end{tabular}
        \end{adjustbox}
\end{sidewaystable}

\end{document}